# A Brain-like Cognitive Process with Shared Methods

Kieran Greer, Distributed Computing Systems, Belfast, UK.
http://distributedcomputingsystems.co.uk
Version 1.4

*Abstract:* This paper describes a new entropy-style of equation that may be useful in a general sense, but can be applied to a cognitive model with related processes. The model is based on the human brain, with automatic and distributed pattern activity. Methods for carrying out the different processes are suggested. The main purpose of this paper is to reaffirm earlier research on different knowledge-based and experience-based clustering techniques. The overall architecture has stayed essentially the same and so it is the localised processes or smaller details that have been updated. For example, a counting mechanism is used slightly differently, to measure a level of 'cohesion' instead of a 'correct' classification, over pattern instances. The introduction of features has further enhanced the architecture and the new entropy-style equation is proposed. While an earlier paper defined three levels of functional requirement, this paper re-defines the levels in a more human vernacular, with higher-level goals described in terms of action-result pairs.

*Index Terms:* Cognitive model, distributed architecture, entropy, neural network, concept tree.

## 1 Introduction

Earlier papers [14][13][10] have described a cognitive model with related processes, based strongly on the human brain. The cognitive model started with a 3-level general design [14] which covers a functionality of: optimisation through link reinforcement, basic aggregations and pattern or node activation through triggers, ultimately leading to some form of thinking. Other papers [11][9][8][7] suggested more localised clustering processes that seem to fit in well with the intentions of that architecture. Two new structures were suggested: the 'concept trees' [8] and a 'symbolic neural network' [11]. Concept Trees are more knowledge-based and could even be used for rote learning. The neural network can work on the symbolic information that the trees create and re-combine it dynamically, even with the help of events. Symbolic refers to any level of concept, or piece of non-numerical information. The main focus of this paper is to further define the pattern creation process and the higher-level aspects of the model, such as creating and using knowledge. If comparisons with the real brain are to be made, then the processes need to be plausible biologically. They do not have to match 100% with current theory however and this paper attempts to put the ideas into context. While the idea of clustering patterns is a general one, vision is more important for this paper, which may be of interest with all of the recent developments in that area.

The rest of the paper is organised as follows: Section 2 gives a recap on the main structures used in the current design. Section 3 briefly introduces some related work. Section 4 describes where the earlier mechanisms fit into the design, especially at the interfaces between the main constructs. Section 5 defines the clustering process more formally through some new equations. Section 6 gives a theoretical example of the clustering process and also a comparison with an existing statistical measurement. Section 7 introduces a new view of the model using higher-level descriptions, while section 8 gives some conclusions to the work.

## 2 Review of the Main Architectures

This section gives some background information on the cognitive model, and the concept trees / neural network model that make up the main architecture.

### 2.1 The Original Cognitive model

A cognitive model was described originally in [14] and is shown again in Figure 1. The design was initially for intelligent information processing, as part of a distributed system such as the Internet. As described in [10]: the first or lowest level allows for basic information retrieval that is optimised through dynamic links. The linking mechanism works by linking nodes that are associated with each other through the interactive use of the system. The second level performs simplistic aggregation or averaging operations over linked nodes, while the third level is more cognitive. It tries to realise more complex concepts autonomously, by linking together associated individual ones. These links form new and distinct entities, and are not just for optimisation purposes. It also attempts to then link the more complex entities, so that a form of thinking can occur. One cluster of linked nodes, when realised, might trigger another cluster and so on. As this flows through different concepts, the network begins to realise things for itself and performs a certain level of thinking. This model is extended in section 7 to be described in terms of higher-level activities that a human might recognise.

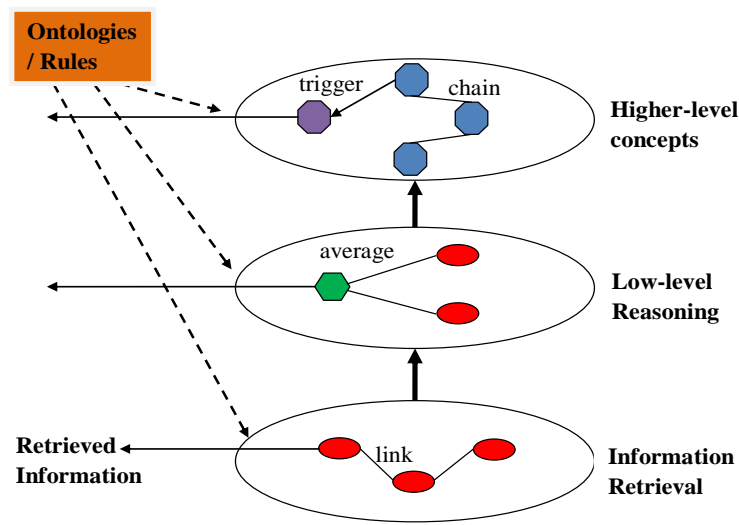

Figure 1. The Original Cognitive Model with new Intelligence Indicators

## 2.2 A More Detailed Upper Level

The architecture of Figure 1 is very general and as already noted, would fit at least two different types of system. If it was ever to be implemented, then the details need to be filled in. More recent work has tried to enhance the theory of the top level [11][9][8][7], which has created the second main construct, consisting of concept trees and a symbolic neural network. This is illustrated in Figure 4 of [7] and Figure 2 below. Concept trees [8] are a knowledge-based structure that can take semistructured or unstructured information and combine it in a consistent way to generate knowledge. They are like a dynamic database and there is a set of rules for constructing them. The symbolic neural network [11] can take the concept trees output and re-group it into other higher-level concept types, relying probably on time-based events. It is symbolic because it groups concepts or symbols more than numbers, but with a time-based element, the information presented would not have to be semantically consistent across all of the links. It could be event-consistent, for example.

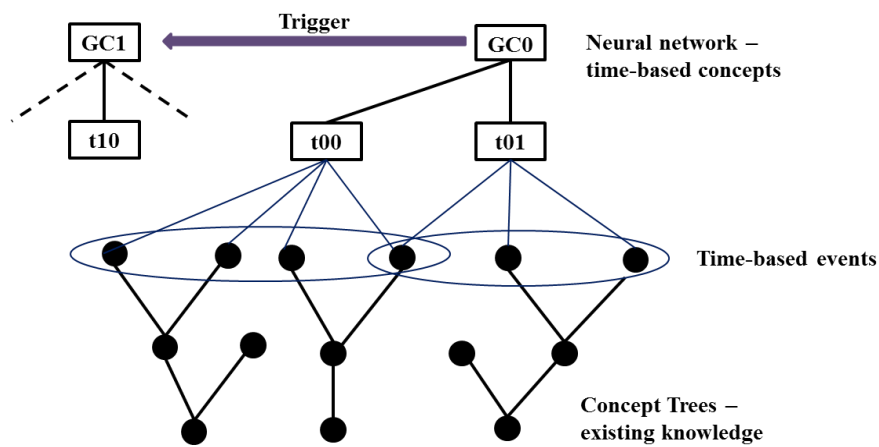

Figure 2. Schematic of the Concept Trees – Symbolic Neural Network architecture.

Each structure broadens from root nodes through branches. One structure branches from the bottom and one from the top and they then meet in a time-based events layer. Searching from the bottom (knowledge-based) requires key nodes to be triggered to activate each tree. The type of indexing provided by that lower level of stimulus has recently been discussed in [18]. Searching from the top (experience-based) is when the acquired knowledge would be tried in different scenarios, to find some type of solution. Therefore, the very top global concepts of the neural network re-combine everything and can also trigger or activate each other, as in the cognitive model of Figure 1. The idea is that a smaller number of more important nodes would trigger the next event.

## 3  Related Work

The model of this paper deals mostly with concepts and therefore might be compared more to shallow (evaluation not complexity) vision systems, than something that performs a deep mathematical calculation. A comparison with Convolutional Neural Networks [5][16][19] is also appropriate and some of those models even overlap the output pattern views, as originally tried in the Neocognitron [5]. The paper [3] is slightly different and uses a probabilistic framework, with various statistical methods for visual object recognition. If using the methods in this paper for the object recognition problem, the idea of a concept would be almost at the pixel level. A single piece of information would be the pixel value itself. The design however allows the concept to be any arbitrary size. The problem that they faced was the classical one of separating individual objects from different scenarios, or when boundaries are not as clear and they note the following requirements and problems:

'First of all, the shape and appearance of the objects that surround us is complex and diverse; therefore, models should be rich (lots of parameters, heterogeneous descriptors). Second, the appearance of objects within a given category may be highly variable, therefore, models should be flexible enough to handle this. Third, in order to handle intra-class variability and occlusion, models should be composed of features, or parts, which are not required to be detected in all instances; the mutual position of these parts constitutes further model information. Fourth, it is difficult, if not impossible, to model class variability using principled a priori techniques; it is best to learn the models from training examples. Fifth, computational efficiency must be kept in mind.'

The flexible structure of this paper's models can help with object categorisation, the use of pattern instances can help with feature identification and the design can also tackle the fourth and fifth problems that they note, as it would be unsupervised and dynamically update itself over time. It is a very different process to their object categorisation, but these keys points still need to be addressed. The paper [29] then maybe shows by how much, recent advances have improved the image classification problem and the Imagenet system [2] also appears to be successful. While those types of system are different to a more general-purpose cognitive model, they do involve filters, feature selection, image overlap and so these types of mechanism are important, both for images or in general. The method used in [29] is to learn specific features (a pattern ensemble) in images, which is again a goal of this paper. This paper's model is therefore still in line with general thinking, even if it is mostly theoretical. Google's DeepMind Learning program [19], for example, was able to learn how to play computer games by itself.

The paper [15] might also be interesting and could tick a few boxes. They produced tests to show that Recurrent Neural Networks are better suited for the Biological Sequencing problem and explain that machine learning techniques have an architectural bias towards different kinds of problem. The Recurrent Neural Networks have an a priori sensitivity to patterns of a sequential nature, with some tolerance to movement and deformation in the patterns. Within that problem domain, there are relatively short sequences known as motifs, or features, and there are also sub-pattern configurations. The problem domain is probably not of interest, but the techniques used and reasons for using them probably is. While the architecture's information flow in this paper indicates feedback, the networks are not really recurrent, but there is still a lot of interactions and self-updating, so maybe it is implicit. The state-to-state occurrence might occur when existing patterns update themselves based on the new pattern input, to achieve the coherence idea (sections 4 and 5), for example. If the symbolic neural network stores each instance that is initially presented and updates those, this again involves large amounts of feedback and self-updating. They then use an error measure of Entropy [22][23], to measure the homogeneity within a group. These ideas have all been mentioned in earlier papers about the current model. Possibly, homogeneity is the correct term instead of cohesion.

The paper [1] is an earlier work on neurophysiological models, where figures 1 and 3 in it illustrate their test theory: The neurons are all connected with each other, which is also a property of the equations in section 5. While they state that it is an unrealistic assumption, the number of synapses involved is still small and it is the basis for their equations. The processes of this paper can maybe add some more direction to those types of connection, but if everything feedbacks to everything else, the pattern size becomes significant. Self-repeating processes also become evident. The paper [18] has suggested that memories can be formed very quickly in the Medial Temporal Lobe (MTL) and Hippocampus, sometimes requiring only one event presentation. This area is in the centre of the brain and so it is not the case that the outer Neocortex, that is thought to carry out intelligent activity, forms all of the memories as well. It would also suggest that information can flow from the brain centre to the outer layer, where it can be processed further. Research by Hawkins and Blakeslee has been mentioned in earlier papers and can be cited again with the tutorial [21], as the work uses similar concepts. The tutorial is a direct attempt to model the Neocortex and includes examples of the computer algorithms. The concept trees might be a form of semantic network and so one reference for that is [24].

## 4 Pattern Clustering and Re-Balancing

The suggestions of this section are again taken from the earlier research and match closely with the detailed model. Pattern clustering and re-balancing requires that firing nodes have something in common. It is recognised however that brain activity can be too much and so while links need to form there is also a controlling mechanism when the signal is reduced. The paper [7] includes an equation intended to model a broad type of interaction between firing patterns and even a very basic type of scheduling. It measures the total excitatory versus the total inhibitory signal for patterns, which is a simplification of an equation in [28]. The counting mechanism [12] has already been considered [11] for creating the initial time-based clusters, or the time-based layer of Figure 4 in [7]. Therefore, if used together, the counting mechanism and the pattern firing equation would be part of the same process. On a technical front, the earlier models created a separate instance of each node for each pattern, whereas this paper allows nodes to be shared between patterns. While the nodes can be shared, they would still require indexed sets of counts, to represent their importance in each individual pattern. So while shared nodes are maybe more realistic, individual pattern instances are also necessary.

### 4.1 Counting Mechanism

A counting mechanism was used in [11] and [12] to try to recognise when an individual node was out of character with a cluster group. For the mechanism, there is an 'individual' count and a 'group' count. The individual count for a node is incremented when that node is presented as part of the group, and the group count for every node is incremented when any node is presented as part of the group. Therefore, the group count continues to be updated, while the local one is updated only when the node is present. This can lead to differences in the values, which can be measured. It was used instead of a basic reinforcement mechanism and was shown to be quite good with noisy input, but not clearly superior. In fact, for the success of the tests [11], it required the addition of a set of automatic rules that were used to determine group count similarities.

It is however, another flexible mechanism and in this paper, it can be used slightly differently, to recognise a type of 'coherence' across a pattern ensemble. In this case, it would not be used to state the exact nodes in the pattern, but to evaluate a level of difference across the pattern. If the counts are different, then maybe there is more than one feature there and the pattern should be split. The counting mechanism can maybe be used to recognise some level of coherence, as coherent nodes should have similar counts. But similarly, incoherent or uncommon nodes groups may be peculiar to a particular pattern and therefore be significant for that reason. When comparing patterns, a feature is something that is the same, but it is also something that is distinctive. If it is the same, then it is indicative of the pattern. If it is different, then it is peculiar to the pattern. There is no default rule here, so this paper presents the general idea and some suggestions for it.

### 4.2 Group Reinforcement

Therefore, to recognise a group of nodes as belonging to a pattern, they should reinforce each other. For this purpose, the equation of [7] section 4.1, is an option and a version of it is used in section 5.2. If used in the current context, the self-reinforcement also takes the size of the pattern into account, which normalises it and might help with convergence or filter processes. As nodes in close proximity are more likely be linked anyway, it is not that bad an idea. The equation of [7] also considers deactivating regions through the inhibitors and so over time, parts of the pattern can either be lost completely or split-up.

Figure 3 illustrates the indexing idea and it is described further in the following sections. On the LHS of the figure there is the input pattern, with a newly created set of linked nodes. There are then 3 existing pattern instances from earlier presentations, on the RHS. The existing patterns have some links that are shared with the new pattern and some other nodes, drawn in blue. The following procedures are a bit descriptive and may not be exact.

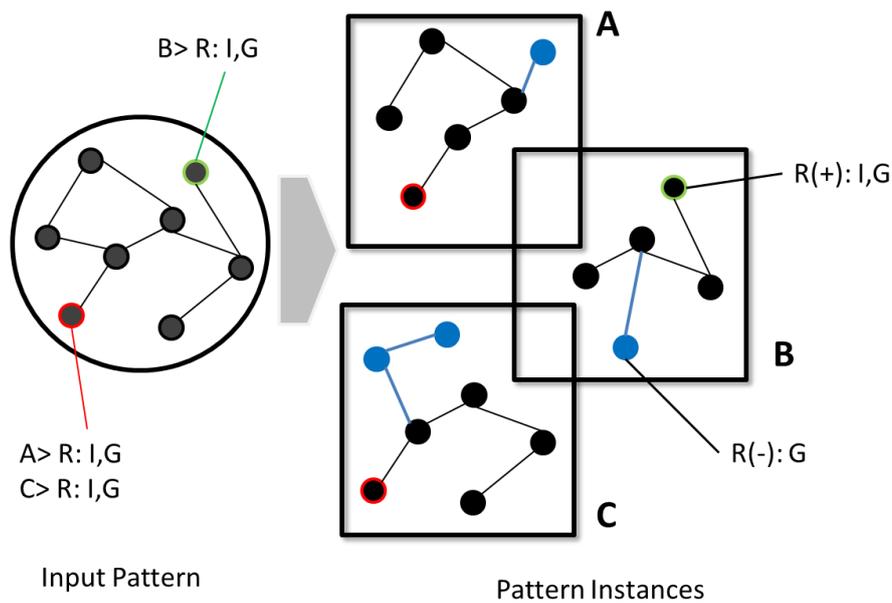

Figure 3. The input pattern nodes update both the individual and group counts for each node in any related pattern instance, with positive reinforcement. The red and green nodes highlight this. The blue nodes only have the group count updated.

For the computer program, when the input pattern reinforces existing ones, it needs to have a separate indexed set of values for every node in it. The earlier research determined that separate instances for unique patterns were required and so unique key sets are a transposition of this. Then for the existing patterns, considering the red highlighted node for example: it is in pattern instances A and C and therefore for them, a reinforcement weight and the individual and group counts can be incremented. This would also be repeated for all shared black nodes in all pattern instances. For any blue nodes however, the individual weight would be decremented and only the group count would be incremented, through the shared node updates. A gap may then build up between the black (shared) node counts and the blue node counts. The blue nodes may then be lost over time, or may get recognised as a more distinctive feature. The same process applies to pattern instance B and the green node. For pattern instance B, it is shown that the green node plus all 3 black nodes have their values updated positively, while the blue node that is not part of the input pattern receives a negative individual update.

### 4.3 ReN

The ReN [9] can help with re-balancing the network, by adding new sub-feature levels when an input signal becomes too strong. It stands for Refined Neuron and is an idea where an input from several neurons can be weakened by adding intermediary ones that need to be energised first. This can also refine the input neuron values, as they become part of a group of signals, thereby providing a fractional value of the eventual output and making the activation process more analogue. It can again happen simply through reinforcement and probably relates to some form of localised re-balancing. So while the ReN can help to automatically build hierarchical structures, the original idea of [14][13] was to make links more accurate by adding a descriptive path to each one, based on metadata or existing knowledge. While that is still important for a distributed system, the idea has been lost a bit in the cognitive model. It is however implicit in the model, where there are lots of hierarchies, representing sub-components, with desired outcomes that may fire to continue the process. So descriptive paths leading to other connections is very much part of the design and would be expected to be part of most distributed, brain-like models.

### 4.4 Cohesion Clustering

Therefore, it might be possible to recognise correct pattern structure through the reinforcement (increment/decrement) of the signals, but also coherence across a pattern. Note that the related work of section 3 mentions the term homogeneity, which might mean the same thing. The process is still very organic, even if a simulator would require storing different sets of count. So while the neuron itself can be shared, for the computer program, storing indexed set of counts for each pattern instance is helpful. If the pattern instances were ever combined, the count sets would have to be re-calculated as part of that.

Figure 4 further illustrates the cohesion problem and solution. The input pattern relates to both of the stored patterns instances. For the smaller one it maps exactly across and so count values would remain consistent. For the larger pattern instance, it maps only to the part shown in red.

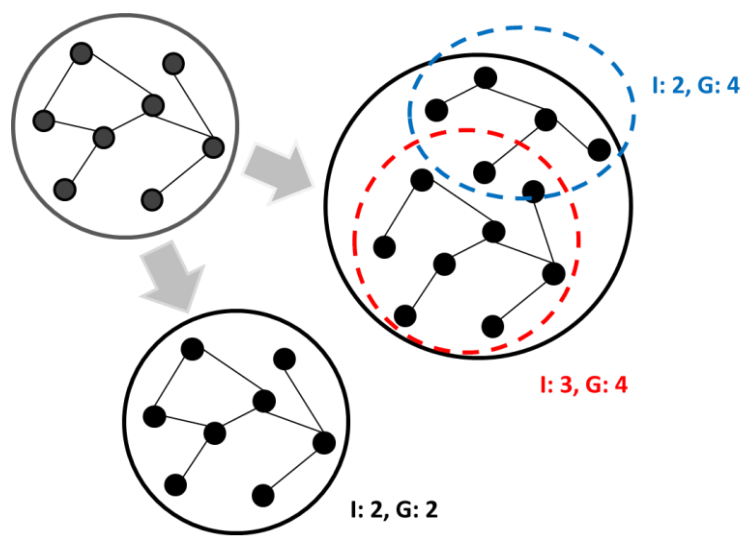

Figure 4. The larger pattern might be made from 2 patterns, where the difference in the individual and group counts can determine this. The smaller pattern is still cohesive and has similar count values across it.

The larger pattern was maybe created earlier and the same would happen if another pattern mapped across to the region shown in blue. The larger pattern therefore develops a difference in the individual and group counts, indicating some level of incoherence. The group count is also a measure of the size of the global pattern and might be useful for normalisation. The system then needs to decide if the pattern should be permanently split, or if a sub-feature has been found. Maybe, as in the real world, a time-based decay would help to separate nodes permanently.

### 4.5 Concept Tree Realisations

The concept tree can therefore be used both as a learned piece of knowledge, or for shallow directional searches and brain models are full of examples of branching structures. If they are to be taken seriously as a bio-inspired structure, the broadening architecture of them needs to be part of a plausible process. It is easy to understand tree structures getting automatically narrower, but to broaden out requires the deliberate addition of new nodes and links to them. Re-balancing is always an option, where excess signal might encourage new neurons to grow, as in a ReN. Or many neuron clusters can interact and link with each other, but still provide specific paths into their own individual set of nodes. An idea of nested patterns might also help. Smaller or less important patterns at the periphery can be linked to by a more common mass in the centre, for example, leading to a kind of tree structure. In which case, it can be less of a deliberate act and more the residual result of a region being stimulated in a particular way [6]. The spatial problem is also helped by the region size.

## 5  Clustering Equations

This section provides some introductory equations that in no way cover all of the required methods, but hopefully show that the system can be built from these basic types of equation. Section 5.1.2 introduces a set of equations based on entropy that may be more generally applicable. The main process is still a reinforcement one, with some count comparisons, where the equations can be divided into ones that are used to create the structure and ones that can be used to subsequently read or activate the structure. It is possible to use either real or binary input signals, but averaging over numbers will still produce a real value for updating counts, for example. Also, a binary input value of 0, for example, can be the same as the node not being present.

## 5.1 To Create the Structure

To create the structure, we can consider either the weight reinforcement or the cohesion counts for each group of nodes, as a pattern ensemble. The counting mechanism is part of the general process and can be considered here, although it might not be the first process applied to the node values. Weight updates can be unit values or proportional to the input signal, for example. The first Equation 1 is the basic reinforcement option. It is still important and can store almost the same information as the counting mechanism. It might be scaled by an individual node weight, for example. Equation 2 is then the counting mechanism individual update for a node and Equation 3 is the counting mechanism group update for a node. Each update would still be based on unique index sets, or a more compact form is described in [6]. Some tests show that the numbers can indeed become confused when all updates are part of the same global pattern:

$$R_{ipt+1} = R_{ipt} +- \omega_I \qquad \qquad \text{Equation 1}$$
$$CI_{ipt+1} = CI_{ipt} + \omega_I \quad \forall \ ((N_i \in IP_t) \wedge (N_i \in P_p)) \qquad \text{Equation 2}$$
$$CG_{ipt+1} = CG_{ipt} + \omega_G \quad \forall \ (N_i \in P_p) \wedge ((N_j \in IP_t) \wedge (N_j \in P_p)) \qquad \text{Equation 3}$$

Where:

$R_{ipt}$ = reinforcement or weight value for node *i* in pattern *p*, at time *t*.
$CI_{ipt}$ = total individual count for node *i* in pattern *p*, at time *t*.
$CG_{ipt}$ = total group count for node *i* in pattern *p*, at time *t*.
$P_p$ = pattern P.
$IP_t$ = input pattern activated at time t.
$N_i$ = Node i.
$\omega_I$ = individual increment value.
$\omega_G$ = group increment value.
$N_g$ = total number of group updates or events.

In words: the individual count is updated for each node every time it occurs in the pattern, as determined by the index set. To represent time-based depreciation, a decrement of all nodes first is possible. The group count update happens every pattern presentation. As it is uniform in this case, it can be updated for the group as a whole and the total number of group updates or events $N_g$ can be useful for factoring. These two equations are essentially the weight update method of the counting mechanism [12]. In fact, with the idea of splitting an existing pattern, the basic reinforcement method of Equation 1 might be best, or the two counts can be used together. For example, reinforcement measures node existence and counting measures similarity.

### 5.1.1 Node Cohesion using Similarity

Comparing the counting mechanism counts can help to determine cohesion across a pattern, by considering each node in the pattern in turn. Therefore, we might measure if a node is cohesive with a pattern using either Equation 4 or Equation 5:

$$Coh_{ip} = ((CG_{ip} - CI_{ip}) / N_G) < \Delta \quad \text{Equation 4}$$

$$Coh_{ip} = (R_{ip} - R_{jp}) < \Delta \quad \text{Equation 5}$$

Where:

$Coh_{ip}$ = true if node *i* is cohesive with respect to the pattern *p* and should be smaller than the allowed difference $\Delta$.
$R_{ip}$ = reinforcement or weight value for node *i* in pattern *p*, compared to node *j*.

In words: cohesion can be a comparison with the upper bound for the pattern, represented by the global count. Alternatively, all nodes with similar weight reinforcement values can be considered together as cohesive.

Therefore, cohesion can indicate pattern membership or it can indicate sub-patterns or features inside of a pattern. If these are larger values, then they maybe represent a positive or indicative feature of the pattern class. If the nodes share smaller values, then they are maybe more specific or peculiar to a sub-type of the pattern. These sub-patterns can actually form a hierarchy of features that fits in well with the other structures of this paper.

### 5.1.2 Pattern Cohesion using Entropy

The last section considered making individual node comparisons. This section describes how it is possible to describe the cohesion equation at a pattern level, by using variances and averages. There are two main considerations for a pattern-level consideration: If the counts are the same for each node in the pattern, then the variance should be very small. The extreme case is when every node is always present and so the count is exactly the same for each node. However, this is not quite enough, because for example, there could be 5 nodes, but each one is updated twice only. In that case the variance would still be very small, but the count comparison would indicate that the nodes were not all updated at the same time. This can be determined by factoring an average individual count by an average global one. Therefore, it might be possible to write the cohesive equation for a whole pattern as the 'variance coefficient (Equation 6) factored by the count difference (Equation 7)'. As with section 5.1.1, the count scaling can come from the local-global count comparison or a combined average reinforcement weight value. Equation 8 is then the pattern cohesion equation itself. The pattern cohesion would evaluate to 1.0 for the best cohesion and 0 for the worst cohesion.

| | |
|---|---|
| Var = (1.0 - (standard deviation / mean local count)) | Equation 6 |
| CFp = (mean local count / mean global count)  or  mean R | Equation 7 |
| Cohp = (Var * CFp) | Equation 8 |

Where:

Var = variance standard deviation that uses the average local count.
$CF_p$ = count difference to scale the variance by.
$Coh_p$ = cohesive value for the whole pattern.

A simplified version for any type of application could probably use some other form of average and total counts. With these equations therefore, a value closer to 1 indicates better cohesion and closer to 0 indicates worse cohesion. It is also good that Equation 8 looks to be related to Entropy [22] more than a distance measure and is therefore about reducing the energy of the system.

### 5.1.3  Optional Cohesive Units

This part is not absolutely necessary but it might be useful. With the idea of features and sub-features, or shared nodes, it might be helpful to be able to update units inside a larger pattern, as a cohesive whole, but separate from the other nodes of the pattern. Also, if a node is currently out of sync with a pattern, it can keep a link with the pattern group, even if the cohesive unit is different. If counts start to match again, the cohesive unit can change back to the pattern, and so on. The equations can get complicated and so they are not included here. A future paper may be able to extend this aspect further.

## 5.2 To Read the Structure

After the patterns have been created and the structure formed, the equation in [7] would be sufficient to calculate an activation value for the pattern as a whole. It was based on an equation in [28] that also includes the sensory input. The sensory input is assumed and so Equation 9 only considers the total excitatory and inhibitory signals, to measure how the patterns will interact with each other. It is also a general reinforcement equation as there is a time element, to state what patterns fire during a time interval, as follows:

$$X_{it} = \sum_{p=1}^{P_i} Ept - (\sum_{k=P_j}^{l} \sum_{y=1}^{m} \sum_{j=1}^{n} (Hjy * \delta))$$  Equation 9

Where, as well as above:

$i \in P_i$ and $not\ j \in P_i$, and
$y \neq t$, and
$X_{it}$ = total input signal for neuron *i* at time *t*.
$E_{pt}$ = total excitatory input signal for neuron *p* in pattern *P*, at time *t*.
$H_{jy}$ = total inhibitory input signal for neuron *j* at time *y*.
$P_i$ = pattern that contains neuron *i*.
δ = weights the inhibitory signal.
P = total number of patterns.
t = time interval for firing neuron.
y = time interval for any other neuron.
n = total number of neurons.
m = total number of time intervals.
l = total number of active patterns.

In words, the equation measures how the activation signal for each pattern changes over the time period. All neurons in the same pattern fire at the same time and send each other their positive signal. Inhibitory input then depreciates the signal and can be obtained from other pattern sources, see [7] for more details. Therefore, while cohesion works inside a single pattern, this equation considers interactions between patterns.

# 6 Example and Comparison

This section gives some worked examples of where the cohesion equations might be used. It also makes a comparison with one of the more established methods called chi-square ([17], for example).

## 6.1 Keywords Example

This example relates to the pattern cohesion equation of section 5.1.2 and can be described using a different technical area of organising text documents. A text management system called Textflo [27] has been developed that includes a document and link Organiser. This organiser is used to group document or link references into categories, where each group can have a list of keywords associated with it. There are 3 categories in total for each group, where the 'Any' category is always added as default. The sub-groups then replace the 'Any' category with their own specific sub-category. For example, a base group 'Artificial Intelligence' with four sub-groups might have the following categories:

*Artificial Intelligence:Any:Any*
*Artificial Intelligence:Cognitive:Any*
*Artificial Intelligence:Cognitive:Pattern*
*Artificial Intelligence:Heuristic:Any*
*Artificial Intelligence:Heuristic:Text*

If the groups that all fall under the same base category name are considered to be related, then their keyword lists should probably be semantically similar. Consider for example, that the set of 5 groups all have keyword lists. Taking the base group as the starting point, a simple count of occurrences might indicate that each keyword used occurs exactly 3 times in total. This would be consistent for the variance, but as there are 5 groups, which would relate to a global count, it still means that keywords are missing from each group. So if this is multiplied by *(3 / 5)* then the coherence value is not as good. If each keyword occurs 5 times however, then it is consistent across all of the groups and the coherence would be the best value of 1.0. This type of calculation would probably be a good indicator that the groups should in fact be listed together. For this example, the keywords list might be a data row and if the keyword is present, it has the value 1 in the row. The rows are then compared with each other and grouped based on the count values.

## 6.2 Pattern Cohesion Example

Another measurement could be to measure if a split pattern has a better composite score than the original one. As a single node will always have a maximum coherence value of 1, splitting a pattern up should not worsen the coherence. This suggests that splitting in general should improve the coherence over each new pattern, but then multiple patterns are less coherent by their very number. For this example, a pattern of 5 nodes with global and local counts is defined in the first table and then worked examples of what the coherence would be after splitting by removing nodes is shown. Note that the average counts need to be re-calculated after the pattern is divided, for each separate part. The initial values are therefore as follows:

Original Pattern with Counts

| gav | 5 | 5 | 5 | 5 | 5 |
|-----|---|---|---|---|---|
| lav | 3 | 3 | 3 | 3 | 3 |
| val | 2 | 4 | 2 | 4 | 3 |

Where:
gav = global mean count.
lav = local mean count.
val = actual local count.

The original pattern cohesion value would then be:

*Nodes 1 – 5:*
gav = 5
lav = 3
Var = Sqrt(Sq(2-3) + Sq (4-3) + Sq (2-3) + Sq (4-3) + Sq (3-3)) / 5 = 0.4
Cohp = 1.0 – (Var / 3) = 1.0 - 0.133  = 0.867 * 3 / 5 = **0.52**

The value is therefore not very cohesive, so try to split by removing a node. In the first example, split by removing node 1:

*Node 1:*
gav = 2
lav = 2
Var = Sqrt(Sq(2-2)) / 1 = 0
Cohp = 1.0 – (Var / 2) = 1.0 – 0  = 1.0 * 2 / 2 = 1.0

*Nodes 2 – 5:*
gav = 4
lav = 3.25
Var = Sqrt(Sq(4-3) + Sq (2-3) + Sq (4-3) + Sq (3-3)) / 4 = 0.43
Cohp = 1.0 – (Var / 3.25) = 1.0 - 0.13  = 0.87 * 3.25 / 4 = **0.707**

In the second example, split by removing node 5:

*Node 5:*
gav = 3
lav = 3
Var = Sqrt(Sq(3-3)) / 1 = 0
Cohp = 1.0 – (Var / 3) = 1.0 – 0  = 1.0 * 3 / 3 = 1.0

*Nodes 1 – 4:*
gav = 4
lav = 3
Var = Sqrt(Sq(2-3) + Sq(4-3) + Sq (2-3) + Sq (4-3)) / 4 = 0.5
Cohp = 1.0 – (Var / 3) = 1.0 - 0.167  = 0.83 * 3 / 4 = **0.623**

Therefore, removing node 1 improves the coherence more than removing node 5. As node 5 is closer to the mean, this should be the case. The node cohesion equations of section 5.1.1 could help to suggest where to split a pattern.

## 6.3   Comparison with Chi-Square using a Benchmark Dataset

This test makes a comparison with the chi-square measurement. The chi-square quantity is commonly used to test whether any given data are well described by some hypothesized function. This can also be called a test for goodness of fit. A typical dataset is not really in the format that has been described in this paper. A node in a pattern does not relate directly to a single variable, but more to a set of values. A sub-feature is more like a subset of data rows, or a separate category, than a subset of variables and so the evaluator would work differently to chi-square, which compares how well sets of variables fit with a hypothesis. The hypothesis is generally formulated from variances and means and so it is similar with respect to the types of value that it uses, but typically compares by adding or removing variables (columns), while the new method compares by adding or removing value sets (rows). Principle Component Analysis is still a very relevant topic, but it is not part of the cohesion equation. In the context of this paper, a pattern node represents a time-based event, or it may be represented by some type of averaged value. For a typical dataset therefore, it may need to be that each data row can represent a single node value; the (sub)patterns would then be the categories that the data rows belong to and the whole dataset can be the largest pattern. Therefore, values are averaged across variables and it is not orthogonal at all. To compare this with chi-square then, the chi-square value for each variable in the data row is added and the average of that sum is used to make the comparison.

The Statlog segment dataset [25][4] is a pattern-recognition dataset and provides a list of image-related values with some category groups. The dataset has been used here to help to show the desired effect of using the new equations. With the dataset, there are rows of variables related to an image feature, for example, centroid region, contrast, RGB. The rows are also categorised as belonging to one of 7 different categories, for example, brickface, sky, foliage. Using the equations of section 5.1.2 in a general sense, a measurement of the whole dataset as a single pattern has been compared with each category being measured separately. This has been done for both the normalised and the not normalised dataset.

|  | **Chi-Square** | **Chi-Square Normalised** | **Cohesion** | **Cohesion Normalised** |
|---|---|---|---|---|
| **Whole Dataset** | 2343.77 | 1677.18 | -3.0E-4 | 1.0E-4 |
| **Category 1** | 59.56% | 17.31% | 313.31% | 1616.71% |
| **Category 2** | 23.48% | 40.23% | 531.46% | 1804.20% |
| **Category 3** | 183.05% | 489.70% | 116.59% | 1393.64% |
| **Category 4** | 98.14% | 53.23% | 524.23% | 1894.73% |
| **Category 5** | 62.20% | 24.51% | 188.57% | 1267.26% |
| **Category 6** | 195.52% | 45.10% | 479.15% | 1864.59% |
| **Category 7** | 78.04% | 29.92% | 301.16% | 1865.49% |

Table 1. Test results of cohesion vs chi-square for normalised and not normalised statlog categories vs the whole dataset. For each category, the value in the table is a percentage of the category result compared with the whole dataset.

The chi-square measurement is variable-based and each category group can have a larger or a smaller value than for the whole group. If the values are not normalised, then the cohesion value was actually negative, but it was positive for the normalised dataset. It is the case however that with the cohesion equation, each category group has a better cohesion value, which is what the theory would suggest. The large percentage differences may suggest that the categories are also quite dissimilar from each other. Therefore, it would be a better metric for this type of problem than chi-square.

## 7  From Optimisations to Activities

This section describes the architecture in terms of more human-oriented activities, instead of an optimisation problem. The automatic processes and methods will be re-worded, to be in terms of a higher-level goal or activity. This is intended to make it easier to conceptualise what each level in the architecture might be able to achieve. There is also some new information or theory, as follows:

### 7.1  Information and Knowledge Are Two Sides of the Same Coin

It is more economical to store correct interactions for a situation than blindly store each situation with a result. The reason is obvious in that a rule can be applied in more than one situation and also the fact that the action is ultimately what is required. Therefore, a compact storage is also more intelligent and puts it in the domain of the physical storage structure. As the brain must economise, it makes sense for it to try to understand something and store that type of information, instead of a blind memory recall of every possible event. Therefore, some type of information transposition has possibly taken place and has the brain been forced to use a more intelligent structure, one that is about interactions between objects, not the objects themselves? At least one definition of intelligence suggests that it can be defined as a better linking structure between nodes, or a better neural efficiency [20][26].

## 7.2 The 'Find – Compare – Analyse' Levels

The cognitive model has now been fleshed out a bit more and some specific methods are defined. It was originally a description of increasing performance (and complexity) that mapped well onto dynamic linking and the manipulation of those links. With the idea that the brain tries to store interactions as much as memories, this increasing complexity could be thought of as going from memory to understanding. The three levels of the cognitive model accommodate this type of transition and can now be put in the context of high-level human-oriented goals or tasks. The levels can be labelled with the functions of 'Find, Compare and Analyse'. While these are the actions of each level, the results of those actions might then be labelled as 'What, Why and How'. To summarise: 'Find' is an initial search process that can be any size. It is looking for a relevant regions and not necessarily specific patterns. 'Compare' can then cross-reference the found regions to identify matching features (or finer granularity). 'Analyse' can then consider the interactions or links between the more specific regions, or patterns, to determine, predict, or even think about them. These fit in with the original cognitive model as illustrated in Figure 5.

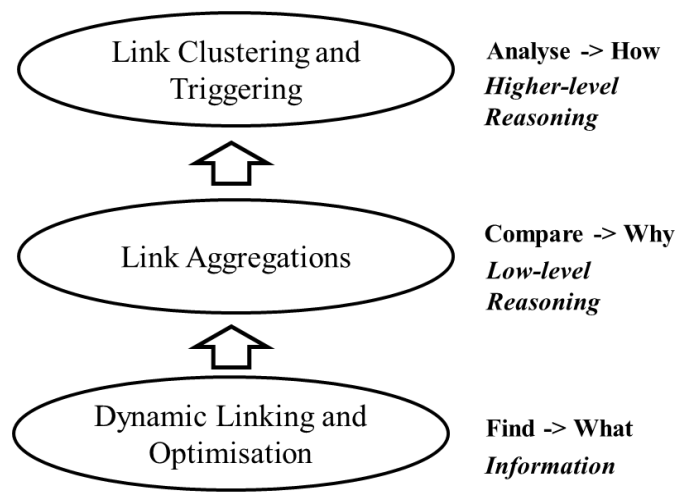

Figure 5. Find-Compare-Analyse as part of the original Cognitive Model

### 7.2.1 Find

Purely as a mechanism, the bottom level of the cognitive model (section 2.1) tries to link-up very efficiently. In higher-level terms this stores information, or in a brain, possibly stores memories. A hierarchical or tree-like structure is also helpful, giving some direction to search operations and adding further definition, such as a sub-pattern. So the initial input stimulus can find or activate broad pattern regions that would contain some level of definition, but can also be refined further. These retrieved patterns might define the 'What' of an input image. If the regions share nodes, then this will necessarily produce a stronger signal in the shared regions. While further refinement is required, this first stage is quite important, because it reduces the number of patterns that are subsequently considered, reducing the problem search space.

### 7.2.2 Compare

The middle level of the cognitive model was defined as an aggregating layer that took lower level links and calculated averaged values for them. In higher-level terms, maybe this aggregation of patterns can result in feature selection. The stronger signals from these shared regions can tell the brain that they make up the essential criteria of the input pattern. If it occurs frequently, then maybe it gets permanently added to a hierarchy or tree structure. Note that the same physical links can be used at different logical levels [9]. This is then 'Why' the input image is being classified as 'What' it is. The ensemble mass resembles something, but on closer inspection, the process can tell why that is the case.

The two types of feature can also be accommodated. A signal can be passed up a feature selector or hierarchy, where higher levels are only activated through a majority stimulus in the lower levels. The larger valued more common features can maybe form at the bottom, leading to smaller valued and less common ones higher up. An automatic system would be able to traverse this type of structure easily. Ideas of a global count can help to produce a percentage estimate of reliability for a feature. Contrary to this however, even very rare features can be significant if they clearly discriminate in certain cases.

### 7.2.3 Analyse

The top level of the cognitive model is then where most of the reasoning takes place. It is probably related more to the behaviour of an object than what it looks like. It is also more time-based than the other two levels and also more dynamic, where static memories or other behaviours can be mixed and processed in different ways. Section 7.1 has suggested that the brain likes to store intelligence and in fact intelligence gets hard-coded in a way similar to memories. The result of this might be a 'How' the object is what it is. Having reasoned about an object's appearance, a person would still confirm the classification through its behaviour. Again, cross-referencing between pieces of behaviour and/or specific features, for example, can subsequently be permanently stored as more knowledge or intelligence, all as part of the same physical structure. After knowing what the image is, what it compares to and how it behaves, we have a good overall understanding of it.

### 7.2.4 Examples

A football and a basketball look quite similar. They are both round and not too different in size. So to tell them apart, more specific features are required, where different patterns on the ball surface can help to distinguish between them. Then, if you were allowed to play with them, trying to kick the basketball would quickly tell you what it was. So while the base features are similar, the higher-level ones tell them apart.

If comparing animals, an Emu and a Robin are not immediately related, but closer inspection tells you that they are both from the Bird family. They are clearly very different sizes and on the surface behave differently, where one flies but the other cannot. However, they both have a round body, round head with a beak, long thin legs with forked feet and of course, feathers. Comparing this with other types of animal makes them more similar than dissimilar. In this case, the common bird features have given enough information to make them more similar than with other animals. Other types of behaviour could also make the two birds similar.

# 8   Conclusions and Future Work

This paper has described a detailed cognitive structure and some processes that appear to be related to each other. One goal of this paper has been to reaffirm earlier research by sharing their structure and methods throughout the cognitive model. A new cohesion measure has also been suggested for determining pattern clusters and the idea of features has been introduced. Introductory equations show how basic methods can be applied and they can all be incorporated into a computer program relatively easily. Some basic tests have shown that the equations would perform as they should, but more detailed tests to determine how accurate they might be, have not yet been carried out. There is still a question about the best use of single, local or global values, but test experiments with the model will help to answer that. The pattern nodes are not orthogonal for this research and that is probably OK, if they are firing in patterns inside of the brain. Possibly, the senses would have more orthogonal input from the outside world.

The 3-level cognitive model was developed for increasing complexity over automatic linking processes. The brain intelligence can further categorise the levels into 'Find, Compare and Analyse'. The intention is again that any processes used would be largely automatic and the results of these actions can be 'What, Why and How'. That is, recognising the patterns, recognising what makes them the same and recognising familiar behaviours in them, which would provide a good basis for 'understanding'. While this is just a new vernacular for the same model, it is also the fact that the methods can be consistently mapped onto the same model and so this gives added confidence for the model as a whole. Hierarchy creation is also about re-organising information and if it can be reactive, then dreaming could be involved. Future work will look at implementing some of the architectural constructs in more detail and try to confirm that they work together in practice.